\documentclass[runningheads]{llncs}

\usepackage[mobile]{eccv}

\usepackage{eccvabbrv}

\usepackage{graphicx}
\usepackage{booktabs}

\usepackage[accsupp]{axessibility}  

\usepackage{verbatim}
\usepackage{amsmath}
\usepackage{amssymb}
\usepackage{bm}
\usepackage{booktabs} 
\usepackage{multirow}
\usepackage{graphicx}
\usepackage{wrapfig}
\usepackage{xcolor} 

\definecolor{mygreen}{RGB}{0,128,0}

\usepackage[pagebackref,breaklinks,colorlinks]{hyperref}

\usepackage{orcidlink}

\begin{document}


\title{DeViDe: Faceted medical knowledge for improved medical vision-language pre-training} 

\titlerunning{Faceted medical knowledge vision-language pre-training}

\author{Haozhe Luo\inst{1,\dagger} \and
Ziyu Zhou\inst{2,\dagger} \and
Corentin Royer\inst{1} \and
Anjany Sekuboyina\inst{1,*} \and
Bjoern Menze\inst{1,*}}

\authorrunning{H. Luo, Z. Zhou et al.}

\institute{Department of Quantitative Biomedicine, University of Zurich, Zurich, Switzerland \and
Shanghai Jiao Tong University, Shanghai, China}

\maketitle

\begin{abstract}

Vision-language pre-training for chest X-rays has made significant strides, primarily by utilizing paired radiographs and radiology reports. However, existing approaches often face challenges in encoding medical knowledge effectively. While radiology reports provide insights into the current disease manifestation, medical definitions (as used by contemporary methods) tend to be overly abstract, creating a gap in knowledge. To address this, we propose DeViDe, a novel transformer-based method that leverages radiographic descriptions from the open web. These descriptions outline general visual characteristics of diseases in radiographs, and when combined with abstract definitions and radiology reports, provide a holistic snapshot of knowledge. DeViDe incorporates three key features for knowledge-augmented vision language alignment: First, a large-language model-based augmentation is employed to homogenise medical knowledge from diverse sources. Second, this knowledge is aligned with image information at various levels of granularity. Third, a novel projection layer is proposed to handle the complexity of aligning each image with multiple descriptions arising in a multi-label setting.  In zero-shot settings, DeViDe performs comparably to fully supervised models on external datasets and achieves state-of-the-art results on three large-scale datasets. Additionally, fine-tuning DeViDe on four downstream tasks and six segmentation tasks showcases its superior performance across data from diverse distributions. 

\keywords{Vision-language Pre-training \and Radiology \and Medical imaging}
\end{abstract}

\section{Introduction}
\label{sec:intro}

Despite the rapid evolution of deep learning models for medical diagnostics, the translation of foundational vision-language pre-training (VLP) models into practical clinical applications remains challenging. The prevailing trend in improving the reliability of VLP in the medical domain largely centres on methodological advancements. Most approaches continue to rely solely on the direct pairing of medical images with raw report data, focusing on refining the model framework to extract and learn from these image-report pairs \cite{chen2023towards, liu2023m, huang2023enhancing, dai2023unichest}, while ignoring other sources of knowledge that offer more granularity. The persisting issue with incorporating an external knowledge base is the data quality (especially in terms of scarcity and noise), although some works have explored techniques to denoise raw text and integrate medical knowledge from multiple sources into the pre-training \cite{zhang2023knowledge,wu2023medklip,huang2023visual,zhang2023text}, they still cannot break through the limitations of coarse granularity of knowledge and reliance on a single source.


The aforementioned limitations in utilizing medical domain knowledge prompt a crucial question: How can we augment medical VLP with medical knowledge from different sources and at different levels of granularity? To answer this question, we propose the DeViDe (\underline{De}finitions and \underline{Vi}sual \underline{De}scriptions), which advances the field by aligning extracted disease-related entities and multi-source knowledge with corresponding visual data. By doing so, DeViDe sidesteps raw reports with complex semantic information, instead extracting medical-related entities combined with enriched domain knowledge. Since traditional VLP models are primarily trained on raw radiological reports, they struggle to encapsulate the disease-specific knowledge necessary for robust diagnostics. To inject detailed disease-specific knowledge for diagnosis, we collected a new dataset that combines radiology-entity terms with their associated visual, radiographic features, providing detailed references for diagnosis.

Our contributions are multifold: (1) We construct a novel dataset of radiology-entity terms along with their associated medical definitions and radiological visual descriptors, providing a complete and direct snapshot of medical knowledge; (2) We present a novel bi-directional transformer architecture capable of processing multiple lengthy medical descriptions. It is trained with multi-granularity losses operating at varying levels of image and text detail, without requiring additional annotations.; (3) we extensively validate our approach across multiple datasets in a zero-shot setting and in fine-tuning, achieving state-of-the-art (SOTA) results compared to five baselines. 

\section{Related Works}
\subsection{Vision-Language Pre-training (VLP)}
Existing VLP methods, which can be categorized into three frameworks, have mainly focused on the development of objectives and architectures to learn multi-modal representations. The first approach is to adopt dual uni-modal encoders~\cite{jia2021scaling, radford2021learning, lu2019vilbert, tiu2022expert} which are composed of separate image and text encoder. CLIP~\cite{radford2021learning} and ALIGN~\cite{jia2021scaling} pre-trained with contrastive learning have been shown to be effective for Image-Text Retrieval (ITR) without object detectors. However, they suffer from performance degradation in other downstream tasks (e.g., Visual Question Answering (VQA), Natural Language for Visual Reasoning (NLVR)). The second approach mainly utilizes a single multi-modal encoder where concatenated text and image representations are used as input. In contrast to the former approach, these works~\cite{chen2020uniter, li2019visualbert, su2019vl, li2020oscar} consistently show promising results on various downstream tasks. However, these methods heavily depend on the pre-trained object detectors which are computationally inefficient. Thus, recent works have tried to replace object detectors with more efficient ones. The last category offsets the shortcomings of the previous approaches by combining them, and achieves state-of-the-art performance. ALBEF~\cite{li2021align} combines them by adding pre-alignment before fusing. Our method is built upon this ALBEF, but deviating from the mainstream VLPs, our attention is on the \textbf{sampling strategy} for efficient pre-training.

\subsection{Medical knowledge enhanced Vision-language Pre-training}
Introducing medical knowledge into VLP model training has demonstrated its effectiveness.  Approaches to incorporating medical knowledge into models can be categorized as either model-based or input-based. Model-based strategies aim to replicate the practices of radiological or diagnostic procedures within the model's design \cite{li2019attention,wang2020learning,huang2020dual,tiu2022expert}. On the other hand, input-based strategies treat medical knowledge as an additional input for computational tasks or as guidance during the model's training process \cite{xie2018knowledge,tan2019expert,zhang2023knowledge,wu2023medklip, melba:2023:003:wittmann}, a technique frequently applied in report generation tasks. Although these methods advance the knowledge-integrated training in VLP, there is still a lack of effective methods for integrating fine-grained discriminative knowledge needed for the diagnostic process.

\section{Method}

Given a dataset with $N$ radiograph-report pairs, $\mathcal{D} = \{(\mathbf{x}, \mathbf{t}) | \mathbf{x} \in \mathcal{X}, \mathbf{t} \in \mathcal{T}$\}, where $\mathbf{x} \in \mathbb{R}^{H\times W}$ denotes a radiograph and $\mathbf{t}$ denotes its corresponding radiology report. As part of our knowledge aggregation (section \ref{sec:knowledge}), we assume the availability of the definitions and visual descriptions for the medical entities related to every sample in $\mathcal{D}$, denoted by $\mathbf{d} = \{d_1, d_2, ... d_k\}$, where $k$ denotes the number of relevant entities per data sample. Our objective is to learn a vision-language model (VLM), denoted by $\mathbf{\Phi}$, which when applied in a zero-shot setting can predict the likelihood of a certain disease as follows: 
$$
\mathbf{\hat{y}} = \mathbf{\Phi}(\mathbf{x}, \small{\texttt{<finding>}}),
$$
where $\mathbf{x}$ denotes a radiograph from the test set and $\mathbf{\hat{y}}$ denotes the likelihood of the presence of a finding \small{\texttt{<finding>}}. In the following sections, we present our solution to address this problem in three stages: first, describing the curation and processing of knowledge, second, the network architecture, and third, the training procedure for aligning the multiple modalities.

\subsection{Knowledge Processing}
\label{sec:knowledge}

\subsubsection{Pre-processing reports.} Radiology reports are extremely noisy and several works have recommended a pre-processing step before their use. For this, we employ RadGraph \cite{jain2021radgraph}, which takes a report ($\mathbf{t}$) as input and recognizes all the radiological entities in it ($e_i | i \in \{1, 2, ... j\}$). It also classifies each of these entities into one of the four categories ($s_i$): anatomy (\small{\texttt{ANAT}}) or observation (\small{\texttt{OBS}}), which can in turn be: \small{\texttt{Definitely Present}}, \small{\texttt{Definitely Absent}}, or \small{\texttt{Uncertain}}. Inspired by \cite{zhang2023knowledge}, we create the final, pre-processed report $\mathbf{e}$ as a concatenation of all the entities and their categories, i.e. $\mathbf{e} = \{e_1, s_i, \small{\texttt{[SEP]}}, e_2, s_2, \small{\texttt{[SEP]}}, ... e_j, s_j\}$. Additionally, we also collect the $M$ most relevant entities in the training dataset into an entity set $\mathcal{E}$, for which the visual descriptions are collected as described below.

\subsubsection{Collecting and processing visual descriptions.}
\begin{figure}[t!]
    \centering
    \includegraphics[width=1\linewidth]{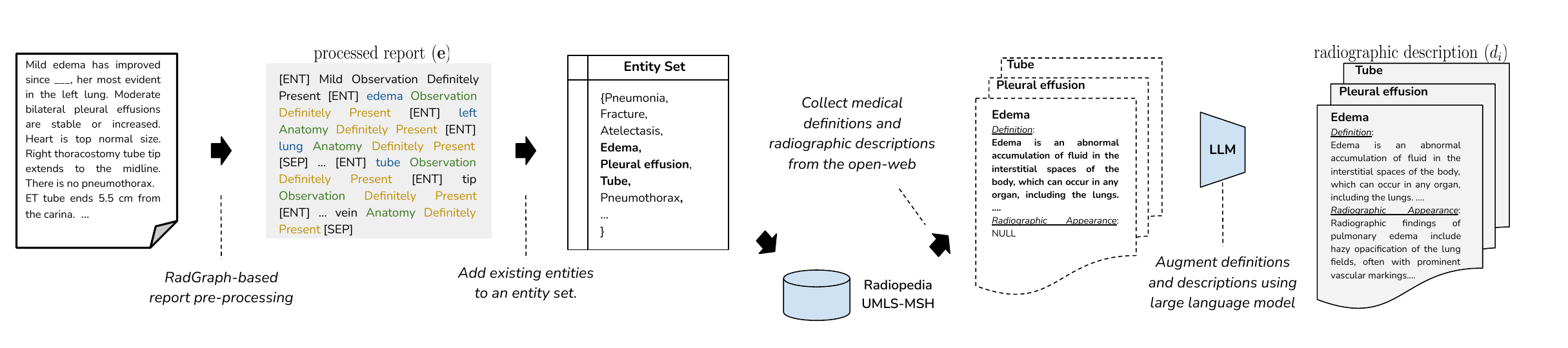}
    \caption{Our knowledge processing pipeline involves preprocessing reports into entities and observations. Entities are queried in the Radiopaedia database for visual descriptions; if none exist, we generate a synthetic description using a Large Language Model (LLM) based on their definitions in a few-shot manner.}
    \label{fig:knowledge}
\end{figure}

Previous works have utilized standardized medical entity descriptions (e.g. radiologist curated definitions or UMLS) as medical knowledge \cite{silva2023foundation, zhang2023knowledge}. On the other hand, trained radiologists refer to established visual descriptions of each finding in order to efficiently diagnose them. These descriptions not only contain fine-grained information about the finding but, often times, discriminative details relative to other entities. Thus, we collect detailed visual descriptions of the entities from an open-source radiology repository (Radiopaedia~\cite{radiopaedia}) and use it to augment our vision-language pre-training. In order to make the collection of this knowledge tractable, we focus only on the entities is $\mathcal{E}$. Specifically, for every entity $e \in \mathcal{E}$, we append the \small{\texttt{MSH}} or \small{\texttt{NCP}} definition (available from UMLS) with the `radiographic features' section of Radiopedia. However, we found that the level of detail in these descriptions was not always consistent. Moreover, not all entities in $\mathcal{E}$ contained this section. To alleviate this challenge, we augment the radiological descriptions using language-based augmentation using a large-language model. We prompt the publicly-available Mixtral-8$\times$7B \cite{jiang2024mixtral} with three examples of disease definitions and radiographic features (for the conditions `atelectasis`, `lung nodule`, and `abscess` because the Radiopaedia data was in the desired format) and then prompt it with a new disease name to generate the relevant text. The few-shot examples help guide the format of the generated data. The prompt is presented as a conversation to the LLM:

\begin{verbatim}
USER: What is a [disease name]? Give me the definition. 
ASSISTANT: Definition: [disease definition] 
USER: What are the radiographic features of [disease name]? 
ASSISTANT: Radiographic features: [disease radiographic features]
\end{verbatim}

This setup can be used to generate any amount of new text quickly. Using LLMs for data augmentation is a common practice that helps improve generalisability. In the medical domain, MAIRA-1 \cite{hyland2023maira} uses it to generate new radiology reports, while LLaVA-Med \cite{li2023llava} generates instruction prompts based on radiology reports. In summary, each training sample in $\mathcal{D}$ is denoted by the triplet $\{\mathbf{x}, \mathbf{e}, \mathbf{d}\}$, where $\mathbf{e}$ denotes the processed report and $\mathbf{d} = \{d_i\}_{i=0}^{k}$ denotes the set of $k$ visual descriptions of the subset of relevant entities present in the training sample. Figure~\ref{fig:knowledge} gives an overview of the text-processing procedure.

\subsection{Network Architecture}
\begin{figure}[t!]
    \centering
    \includegraphics[width=12.5cm]{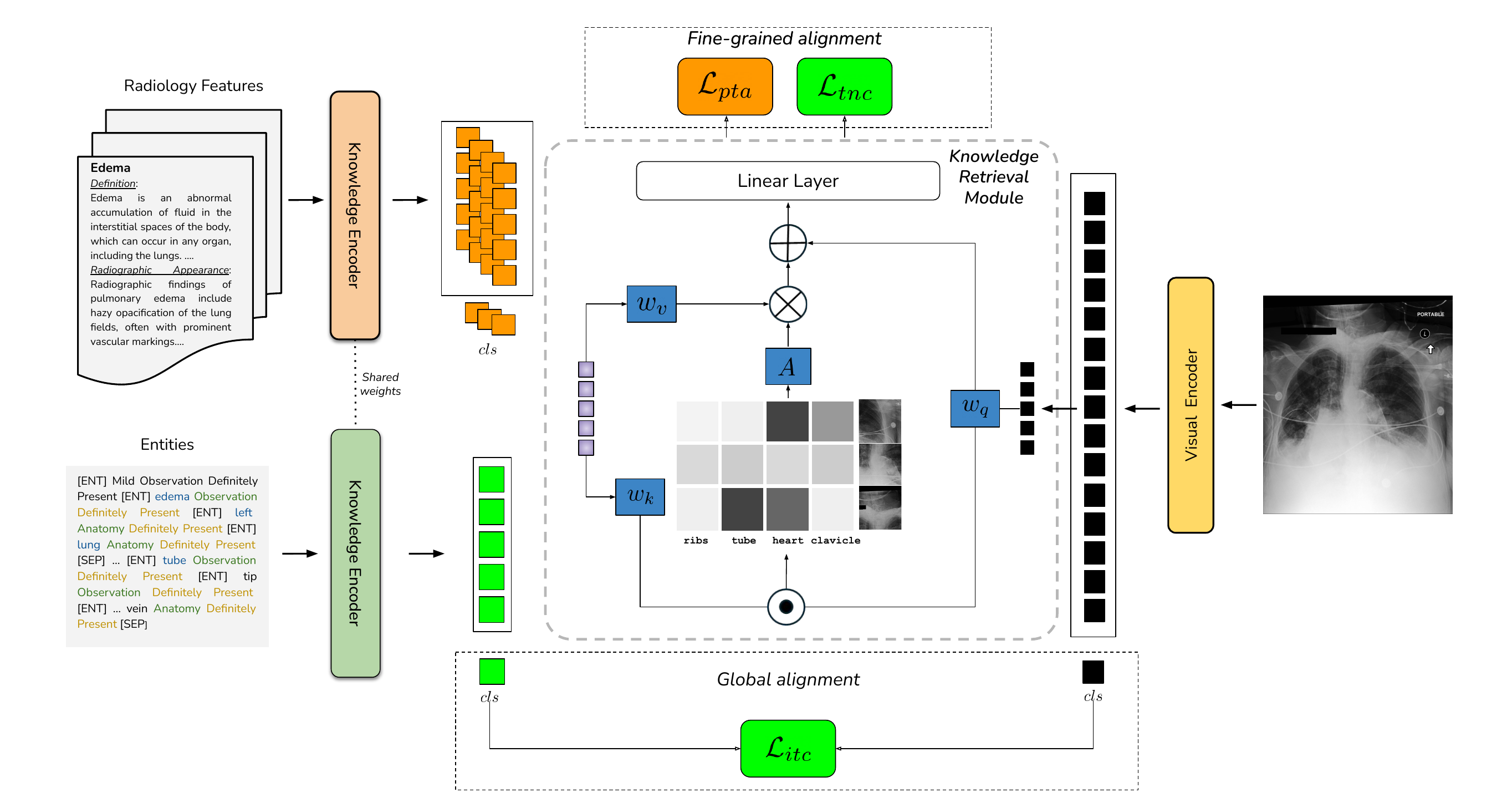}
    \caption{The proposed DeViDe framework includes: encoding the image using a visual encoder (Sec.~\ref{sec:image_enc}), entity knowledge and visual descriptors with a knowledge encoder (Sec.~\ref{sec:knowledge}), capturing image-level entities-to-image correspondence using ITC  loss, and employing Transformer-based Fusion layers for fine-grained alignment between image patches and visual descriptor tokens using the TNC and PTA losses (Sec.~\ref{sec:losses})}
    \label{fig:framework}
\end{figure}

Our model consists of two encoding streams, one for images and one shared between the pre-processed report and the radiographic descriptions. Both the image and text facets are fused into one joint feature space using a specialized cross-view fusion module and then decoded into logits resulting in likelihoods. Fig.~\ref{fig:framework} illustrates an overview of DeviDe's architecture.

\subsubsection{Image Encoding.}
\label{sec:image_enc}

We encode a radiograph using a visual backbone $\Phi_\text{image}$, such as:
\begin{equation}
v = \Phi_{\text{image}}(\mathbf{x}) \in \mathbb{R}^{P \times d},
\end{equation}

where, when utilizing the vision transformer (ViT, \cite{dosovitskiy2020image}), $P$ and $d$ denote the number of patches into which the image is split and $d$ denotes the feature dimension for every patch. We directly employ the last feature block of ViT-B, allowing for a richer feature representation, and leveraging the global and local contextual information. We denote ViT's learnable \texttt{class} token by $v^{cls} \in \mathbb{R}^d$, which represents the global encoding of $\mathbf{x}$.


\subsubsection{Report and Knowledge Encoding.}
\label{sec:knowledge_enc}

The processed report ($\mathbf{e}$) and the set of relevant radiographic descriptions ($\mathbf{d}$) are encoded using Med-KEBERT\cite{zhang2023knowledge} as a pre-trained text encoder. We first tokenize $e$ and $\{d_{i}\}_{i=0}^k$ and then encode them into as text embeddings. For aggregating the information from the $k$ documents, we propose a learnable linear layer initialized with the category of the presence or absence of the entity corresponding to the description $d_i$. Specifically, the weight is set at 1, 0, or -1 if the entity is Definitely Present, Uncertain, or Definitely Absent. This is done so as to allow for the visual information of both the positive and negative classes to affect the training process. The report and description encoding is represented as follows: 


\begin{equation}
t_e = \Phi_{\text{text}}(\mathbf{e}) \in \mathbb{R}^{T \times d}
\end{equation}
\begin{equation}
t_d =  \sum_{d_i \in \mathbf{d}} w_i\Phi_{\text{text}}(d_i) \in \mathbb{R}^{T \times d}
\end{equation}

where $T$ denotes the number of text tokens, $d$ denotes the feature dimension, and $w_i$ denotes the learnable weight of the $i_{th}$ description embedding. Note that textual knowledge encoder also outputs the global \texttt{CLS} tokens, $t_{e}^{cls}$ and $t_{d}^{cls}$ , which contain the overall information of the reports and the descriptions.




\subsection{Training}
\label{sec:losses}

In this section, we describe the training procedure of DeViDe which aligns the image and text embeddings at varying granularity. This departs from prior works that typically align the image and text embeddings using contrastive loss.

\subsubsection{Global radiograph-report alignment}

Global feature alignment between the image and the radiology report is achieved using the ubiquitous image-text contrastive loss ($\mathcal{L}_{itc}$), which maximizes the mutual information between the image and report representations. For this loss, we use the report embedding, which describes the overall image. Given a training radiograph-report-description triplet $(\mathbf{x}, \mathbf{e}, \mathbf{d})$, the learning objective is to maximize the similarity between the embedded image and report, specifically their \texttt{CLS} token. Formally,

\begin{equation}
\mathcal{L}_{itc} = -\log \frac{\exp\big(\text{sim}(v^{cls}, t_e^{cls}) / \tau\big)}{\sum_{t_e^{'cls} \in B} \exp\big(\text{sim}(v^{cls}, t_e^{'cls}) / \tau\big)},
\label{eq:l_itc}
\end{equation}

where, $\text{sim}(\cdot, \cdot)$ denotes the cosine similarity score, $\tau$ denotes the temperature scaling used for adjusting the sharpness of the distribution, and $B$ denotes the current training batch.

\subsubsection{Fine-grained alignment using cross-modality fusion}
In contrast to natural images, the information in medical images is fine-grained. More so when we are concerned about specific sub-domains, e.g. chest radiographs. In order to amplify the model's discriminating ability on difficult cases, we choose to co-align the tokens of the radiograph with those of the report and the radiographic description using a token-level cross-attention module termed the \emph{knowledge retrieval module} (KRM). The KRM contains multiple cross-attention layers where the key and value pairs are projections of either the report tokens or the description tokens which are queried by the visual embeddings corresponding to local patches.

Given a visual query \( v\in \mathbb{R}^{P \times d} \), it is initially encoded as \(Q\) by a projection layer, while the text embedding \( t\in \mathbb{R}^{T \times d} \) is encoded as \(K\) and \(V\). Note that $t$ could denote either $t_e$ or $t_d$. The visual query $Q$ interacts with text embeddings \(K\) and \(V\) to facilitate a fine-grained matching by computing how an image patch attends to a text token,
\begin{equation}
v_{pa} = \text{softmax}\left(\frac{QK^T}{\sqrt{d}}\right)V,
\end{equation}
where $v_{pa} \in \mathbb{R}^{P\times d}$ represent the text-attended, enriched visual embeddings. 

\paragraph{Tough negative contrastive loss} is computed to enhance the model's ability to discriminate difficult cases, inspired by metric learning. Specifically, the model is tasked to distinguish between correctly-aligned image-text pairs and closely analogous, but non-matching pairs. The learning objective for the directional alignment of image-to-text can be formalized as: 

\begin{equation}
\mathcal{L}_{tnc}(v_{pa}, t_e) = -\log\left(\frac{\exp\big(s(v_{pa}, t_e)\big)}{\exp\big(s(v_{pa}, t_e)\big) + \exp\big(s(v_{pa}, \bar{t}_e\big)}\right)
\label{eq:l_tnc}
\end{equation},

where $\bar{t}_e$ denotes the hard-negative example within the batch $B$ as measured by cosine similarity along the text embeddings. The scalar alignment score $s(v_{pa}, t_e)$ is computed directly from the enriched image embeddings by pooling them along the patch-dimension (P) and projecting them using a learnable, linear layer of dimension $\mathbb{R}^{1\times d}$. Similarly, $\mathcal{L}_{tnc}(t_e, v_{pa})$ denotes the loss computed in the text-to-image direction, with negative mining in the image embedding space.

\paragraph{Local radiograph-description alignment} is performed by tasking the model to match image patches with specific details in their radiographic description. This is done by minimizing the following cross-entropy loss aimed at patch-text alignment based on the alignment score computed as described above, albeit with a separate projection layer. Formally,

\begin{equation}
\mathcal{L}_{pta} = -\log\left(\frac{\exp\big(s(v_{pa}, t_d)\big)}{\sum_{t_d' \in B} \exp\big(s(v_{pa}, t_d')\big)}\right)
\label{eq:l_pta}
\end{equation}


\subsubsection{Overall loss} 
The overall training objective is obtained by combining the global image-text contrastive loss (Eq.~\ref{eq:l_itc}), the fine-grained tough-negative loss (Eq.~\ref{eq:l_tnc}), and the local patch-text alignment of the radiographs and the descriptions (Eq.~\ref{eq:l_pta}). Additionally, we support the training by employing a weakly-supervising, binary cross-entropy loss ($\mathcal{L}_{bce}$) if queries from $\mathcal{E}$ appear in the report. Specifically, the names of the entities are encoded by $\\Phi_{text}$ and refined using the knowledge-retrieval module to produce the output logits. The total loss to be minimized is. 

\begin{equation}  \label{eq: total loss}
    \mathcal{L} = \mathcal{L}_{bce} + \mathcal{L}_{itc} + \mathcal{L}_{tnc} + \alpha\mathcal{L}_{pta},
\end{equation}
where $\alpha$ regulates the contribution of radiographic descriptions to the training process.

\subsection{Inference}
During the inference phase, for a given test image \(X\), the model is designed to identify specific entities or diseases. Given an entity query set \(q\) and the test image \(X\), the process begins with encoding these inputs to generate embeddings \(t_q\) and  \(v\), respectively. Specifically, \(t_q\) is generated as the query embedding, while \(v\) is utilized as both the key and value embeddings. These embeddings are then fed into a knowledge retrieval module, which employs entity-specific visual markers to refine the query embedding, resulting in \(t_q^{\text{refined}}\). Subsequently, an MLP layer maps the feature dimension of \(t_q^{\text{refined}}\) from \(d\) to 1, producing an output that indicates the presence or absence of each queried entity.


\section{Implementation details}

\subsubsection{Pretraining.}
With ViT-B as the visual backbone and Med-KEBERT as the textual backbone, we pre-train on the MIMIC-CXRv2 dataset \cite{johnson2019mimic} on an image size of 224. We utilize the AdamW optimizer with learning rates \( lr = 5 \times 10^{-5} \) and \( lr_{\text{warmup}} = 1 \times 10^{-6} \). We optimize on A100 80G GPUS with a total batch size of 128 for a total of 100 epochs with the first 25 epochs for warm-up. DeViDe, pre-trained on MIMIC-CXR, is used out-of-box in a zero-shot setting for disease classification on ChestX-ray14~\cite{wang2017chestx}, CheXpert\cite{irvin2019chexpert} and PadChest\cite{bustos2020padchest} datasets.   

\subsubsection{Finetuning.}
The pre-trained DeViDe can also be fine-tuned for downstream classification and segmentation tasks to evaluate its generalizability. For classification tasks, we append a randomly-initialized, linear layer to the output. Specifically, to the CLS token output for ViT-B based models and to the average-pooled feature maps from the last layer for ResNet-based models (evaluated as part of ablation). The area under the ROC curve (AUC), Matthew's Correlation Coefficient (MCC), accuracy (ACC), and F1 metrics are used to evaluate the multi-label classification performance on NIH ChestX-ray14~\cite{wang2017chestx}, CheXpert\cite{irvin2019chexpert} and Shenzhen CXR~\cite{jaeger2014two}). Accuracy is used to assess the multi-class classification performance (RSNA Pneumonia~\cite{rsna}). For the segmentation task, we use UperNet\cite{xiao2018unified} as base model, replacing our pre-trained ViT-B with a randomly initialized mask prediction head and fine-tuning it on four datasets:  JSRT~\cite{shiraishi2000development}, ChestX-Det~\cite{lian2021structure}, SIIM-ACR~\cite{siim-acr}, and Montgomery~\cite{jaeger2014two}. In the JSRT dataset, we independently train three models, one for each of lung, heart, and clavicle. Dice is used to evaluate the segmentation performance. For fine-tuning the segmentation model, we use the AdamW optimizer with a cosine learning rate scheduler, linear warm-up of 20 epochs, and a total of 150 epochs. We employ a maximum learning rate of 0.0005 over a batch size of 64 images each of 224. We train on a single V100 16G GPU.

\subsubsection{Baselines}
We evaluate our approach against a range of existing state-of-the-art (SOTA) methods for medical VLP, including ConVIRT \cite{zhang2022contrastive}, GLORIA \cite{huang2021gloria}, BioViL \cite{boecking2022making}, CheXzero \cite{tiu2022expert}, Medklip \cite{wu2023medklip}, and KAD \cite{zhang2023knowledge}, focusing on zero-shot inference tasks. For fine-tuning,  we also benchmark our model against the aforementioned methods where applicable as well as on models trained from scratch or those pre-trained on the ImageNet.

\section{Results}

\subsection{Zero-shot setting}


\begin{figure}[!t]
    \begin{minipage}{0.5\textwidth}
        \includegraphics[width=\linewidth]{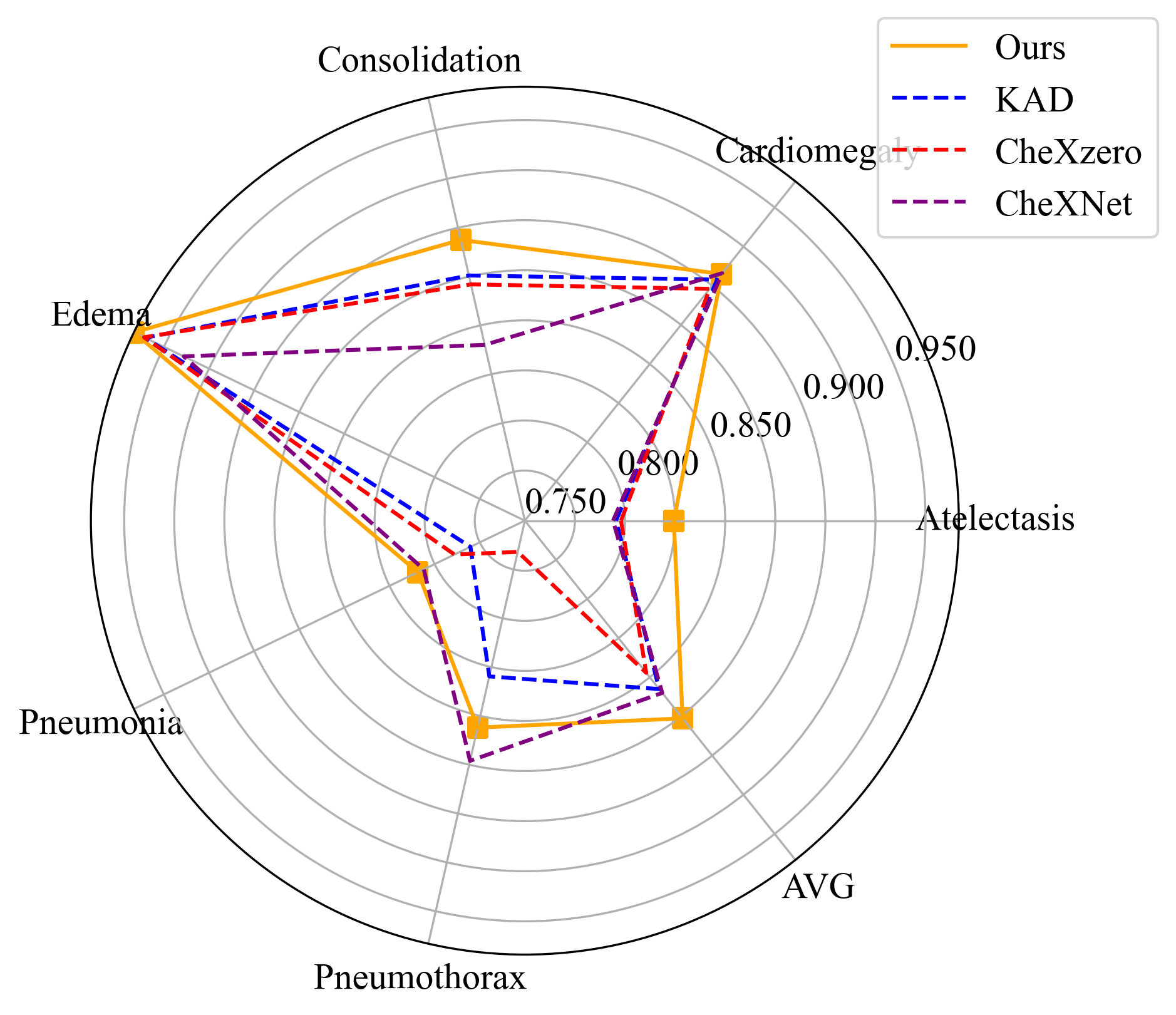}
    \end{minipage}%
    \hfill
    \begin{minipage}{0.45\textwidth}
        \captionof{figure}{The radar chart showcases our method's AUC scores for seven PadChest diseases, demonstrating superior performance compared to previous state-of-the-art. Notably, our approach surpasses CheXNet's fully-supervised performance despite being evaluated in a zero-shot setting.}
    \label{fig:radarpadchest}
    \end{minipage}    
\end{figure}

\begin{figure}[t!]
    \centering
    \includegraphics[width=\linewidth]{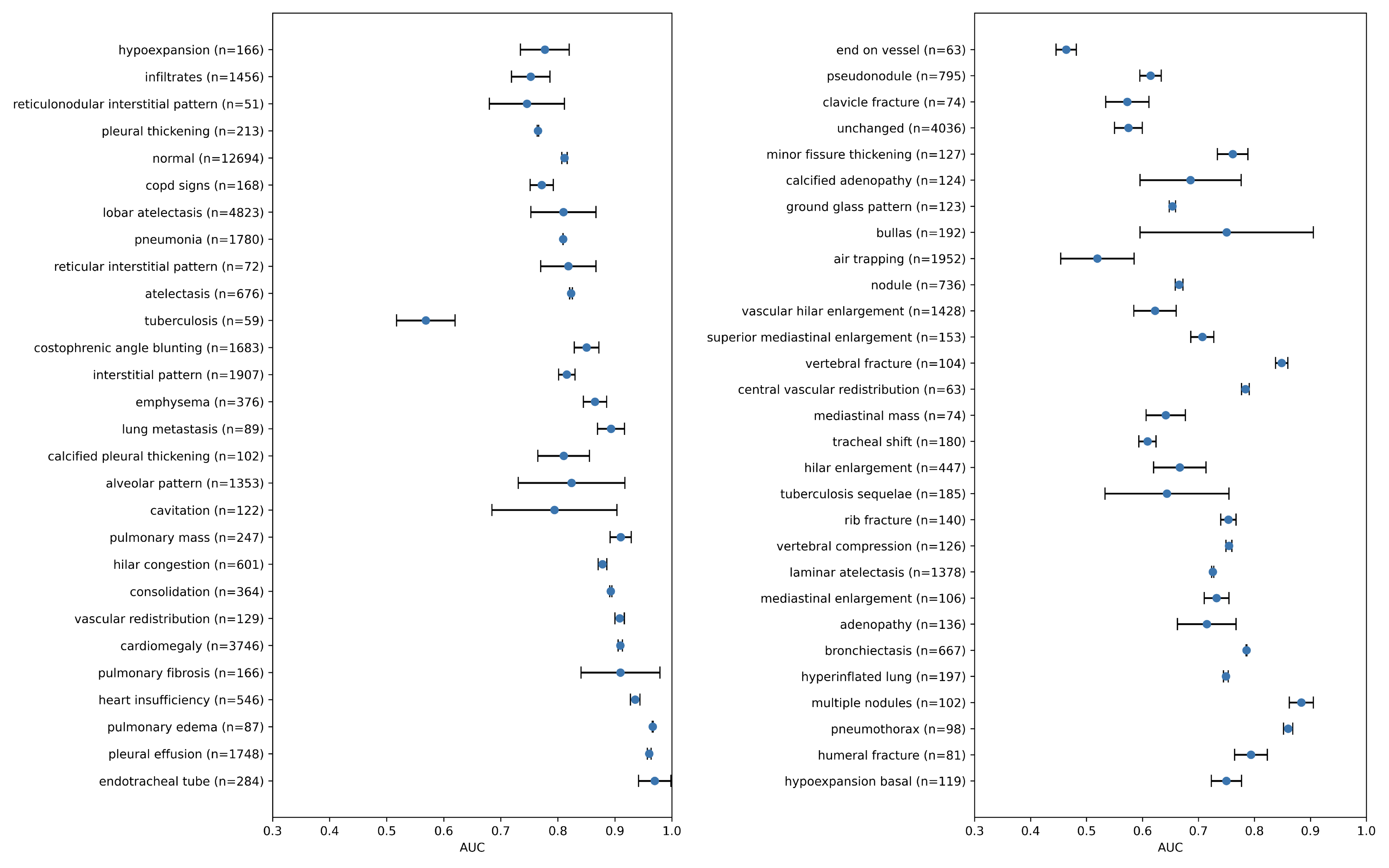}
    \caption{In the PadChest dataset, we evaluated radiographic findings deemed high-importance by a radiologist, each with a sample size exceeding 50, presenting their mean AUC and 95\% confidence intervals (CI). To ensure generalization, we externally validated our model on a subset of 39,053 human-annotated chest X-rays from PadChest, with no labeled samples from this set used during training. DeViDe achieved an AUC of at least 0.900 for eight findings and at least 0.700 for 43 out of 57 findings.}
    \label{fig:padchest_open}
\end{figure}

In this section, we evaluate our method in a zero-shot learning (ZSL) setting and compare it to relevant baselines over three three prominent datasets, as recorded in Table \ref{tab:results}. The classes evaluated in ChestXray14 and CheXpert datasets are mostly included in MIMIC-CXR and can been treated as a closed-world ZSL setup. Observe that our model outperforms all baselines methods on all metrics (except ACC). Notably, we see substantial improvement in diseases containing detailed radiographic descriptions, e.g. for Atelectasis and Edema, our model increased the AUC score by about 6\%, from \textbf{0.81} to \textbf{0.87} and from \textbf{0.90} to \textbf{0.96}, respectively, surpassing CheXzero. Likewise, AUC for pleural effusion improved from \textbf{0.93} to \textbf{0.96}. When evaluating on PadChest, our model achieves superior performance in \textbf{5} out of \textbf{6} unseen classes listed in CheXNet, inspite of CheXNet being fully-supervised. Our model also demonstrates better ZSL performance on other pre-trained models, as illustrated in Fig.~\ref{fig:radarpadchest}. Furthermore, our model demonstrates robust performance, achieving an AUC of at least \textbf{0.90 in eight findings} and at least \textbf{0.70 in 43 out of 56 radiographic findings} within the PadChest dataset. This showcases our model's effectiveness in open-world disease detection in a dataset with a large domain shift. Detailed, class-wise performance is presented in Fig.~\ref{fig:padchest_open}.

The ZSL performance of DeViDe highlights its robustness in handling the inherent variability in multi-center datasets, particularly with diverse range of medical entities. The high performance on open-world classes   can be attributed to the transferable knowledge learnt from from radiographic descriptions during the pre-training phase.


\begin{table}[!t]
\centering
\caption{Performance comparison against SOTA methods in zero-shot classification. The macro average of scores on all diseases are reported.}
\label{tab:results}
\setlength{\tabcolsep}{4pt} 
\begin{tabular}{lccccccc}
\toprule
\multirow{2}*{Methods Dataset } & \multicolumn{3}{c}{ChestX-ray14} & \multicolumn{3}{c}{Chexpert} & \multicolumn{1}{c}{PadChest} \\
\cmidrule(r){2-4} \cmidrule(l){5-7} \cmidrule(l){8-8}
                        & AUC & F1 & ACC & AUC & F1 & MCC & AUC \\
\midrule
ConVIRT                 & 0.560        & 0.135       & 0.459        & 0.590         & 0.264            & 0.231          & - \\
GLoRIA                  & 0.610        & 0.174      & 0.503       & 0.750         & 0.570            &  0.501         & - \\
BioVil                  & 0.662      & 0.192       & 0.633      & 0.693         &  0.463            &  0.368         & 0.847 \\
CheXzero                & 0.730        & 0.214       & \textbf{0.828}        & 0.889         & 0.606           & 0.523          & - \\
MedKlip                 & 0.726        & 0.244       & 0.796       & 0.879         &0.614            & 0.540          & - \\
KAD                     & 0.768         & 0.312            & 0.820             & 0.871         & 0.593            & 0.510         &0.858 \\
\textbf{Ours}           & \textbf{0.777}  & \textbf{0.315}            & 0.823             & \textbf{0.900} & \textbf{0.640}             & \textbf{0.566} & \textbf{0.876} \\
\bottomrule
\end{tabular}
\end{table}

\subsection{Fine-tuning Evaluation}
In this part, we start with a pre-trained model as the base and proceeding to train it comprehensively on the downstream tasks of classification and segmentation. We compare our method's performance to that trained from scratch in a fully-supervised manner, as well as to that fine-tuned from other pre-training approaches including the standard ImageNet initialization. 

\begin{table}[!t]
        \centering
        \setlength{\belowcaptionskip}{0.1cm}
        \caption{Comparison with fully-supervised pre-trained models on different classification tasks via AUC metric.}
        \label{tab: fully-supervised-seg} 
        \begin{tabular}{cccccc}
        \toprule    
             \multirow{2}*{Backbone} & Pretraining data & \multirow{2}*{ChestX-ray14} & \multirow{2}*{CheXpert} & \multirow{2}*{ShenZhen} & RSNA  \\
              & and methods& & & & Pneumonia \\
        \midrule
            \multirow{3}*{ResNet-50} & Random & 80.40 $\pm$ 0.05 & 86.60 $\pm$ 0.17 & 90.49 $\pm$ 1.16 & 70.00 $\pm$ 0.50 \\
            & ImageNet-1K & 81.70 $\pm$ 0.15 & 87.17 $\pm$ 0.22 & 94.96 $\pm$ 1.19 & 73.04 $\pm$ 0.35\\
            & \textbf{DeViDe} & \textbf{82.29 $\pm$ 0.11} & \textbf{88.81 $\pm$ 0.45}   & \textbf{97.94 $\pm$ 0.20} & \textbf{74.45 $\pm$ 0.10}\\
        \midrule
            \multirow{3}*{ViT-B}& Random & 70.84 $\pm$ 0.19 & 80.78 $\pm$ 0.13 & 84.46 $\pm$ 1.65 & 66.59 $\pm$ 0.39\\
            & ImageNet-1K & 77.55 $\pm$ 1.82 & 83.32 $\pm$ 0.69 & 91.85 $\pm$ 3.40 & 71.50 $\pm$ 0.52\\
            & \textbf{DeViDe} & \textbf{81.32 $\pm$ 0.08}& \textbf{88.33 $\pm$ 0.27}   & \textbf{98.03 $\pm$ 0.24} & \textbf{74.57 $\pm$ 0.42}\\
        \bottomrule
        \end{tabular}
    \end{table}

\begin{table}[!t]
    \centering
    \setlength{\abovecaptionskip}{0.1cm}
    \setlength{\belowcaptionskip}{0.1cm}
    \caption{Comparison with other pre-training methods on AUC score. Different Models are end-to-end fine-tuned on the ChestX-ray14 Dataset.}
    \label{tab:fully_supervised_sota} 
    \begin{tabular}{cccccccc}
    \toprule
        Pre-training data & \multicolumn{6}{c}{Methods} \\
        \cline{2-8} 
        and methods & ConVIRT & GLoRA & BioVil & MedKlip & KAD & PEAC\cite{zhou2023learning} & \textbf{DeViDe} \\
    \midrule
        ChestX-ray14 & 0.808 & 0.800  & 0.800 &0.801 & 0.805 & 0.800 & \textbf{0.813} \\
    \bottomrule
    \end{tabular}
\end{table}

\subsubsection{Classification}
We conduct experiments on four distinct datasets, utilizing fractions of the data for the fine-tuning phase, in accordance with established precedents in the literature. As depicted in Table.~\ref{tab: fully-supervised}, fine-tuning DeViDe demonstrated significant enhancements in performance over models trained from scratch with random initialization as well as with ImageNet initialization. The performance was consistent across four distinct medical imaging datasets with both ResNet-50 and ViT-B encoders. Notably, the Vision Transformer derived greater benefits from the DeViDe pre-training, as evidenced by at least a \textbf{4\%} increase in AUC scores on substantial datasets such as ChestX-ray14 and CheXpert, relative to ImageNet initialization. On the Shenzen and RSNA Pneumonia datasets, which are smaller in scale, the benefits of our pre-training are more magnified, reaching up to a \textbf{6.2\%} improvement in AUC. This shows the strong feature reuse capability of our model. 

In Table.~\ref{tab:fully_supervised_sota}, we fine-tune our method as well as the other baselines on the ChestX-ray14 dataset. Our model demonstrates significantly better performance compared to the state-of-art. Despite KAD utilizing medical knowledge from UMLS and MedKLIP employing curated medical definitions, DeViDe's superior AUC underscores the contribution of the granular, radiographic descriptions of the entities during the pre-training.

\subsubsection{Segmentation}
 With regards to fine-tuning a segmentation model, we widely validate the dense prediction performance on JSRT~\cite{shiraishi2000development} for multi-organ segmentation, ChestX-Det~\cite{lian2021structure} for multi-diseases conditioned segmentation, SIIM-ACR~\cite{siim-acr} for Pneumothorax segmentation and Montgomery~\cite{jaeger2014two} for Effusions and Miliary patterns. We transfer the pre-trained encoder, along with a randomly-initialized decoder, to each target task through end-to-end fine-tuning of all parameters. In Table.~\ref{tab: fully-supervised}, we present the performance of fine-tuning based on DeViDe and compare it with that of full training from random initialization as well as ImageNet initialization. Notice that our method demonstrates an stable improvement in Dice score on different data distributions, albeit small. 

\begin{figure}[!t]
    \centering

    \includegraphics[width=\linewidth]{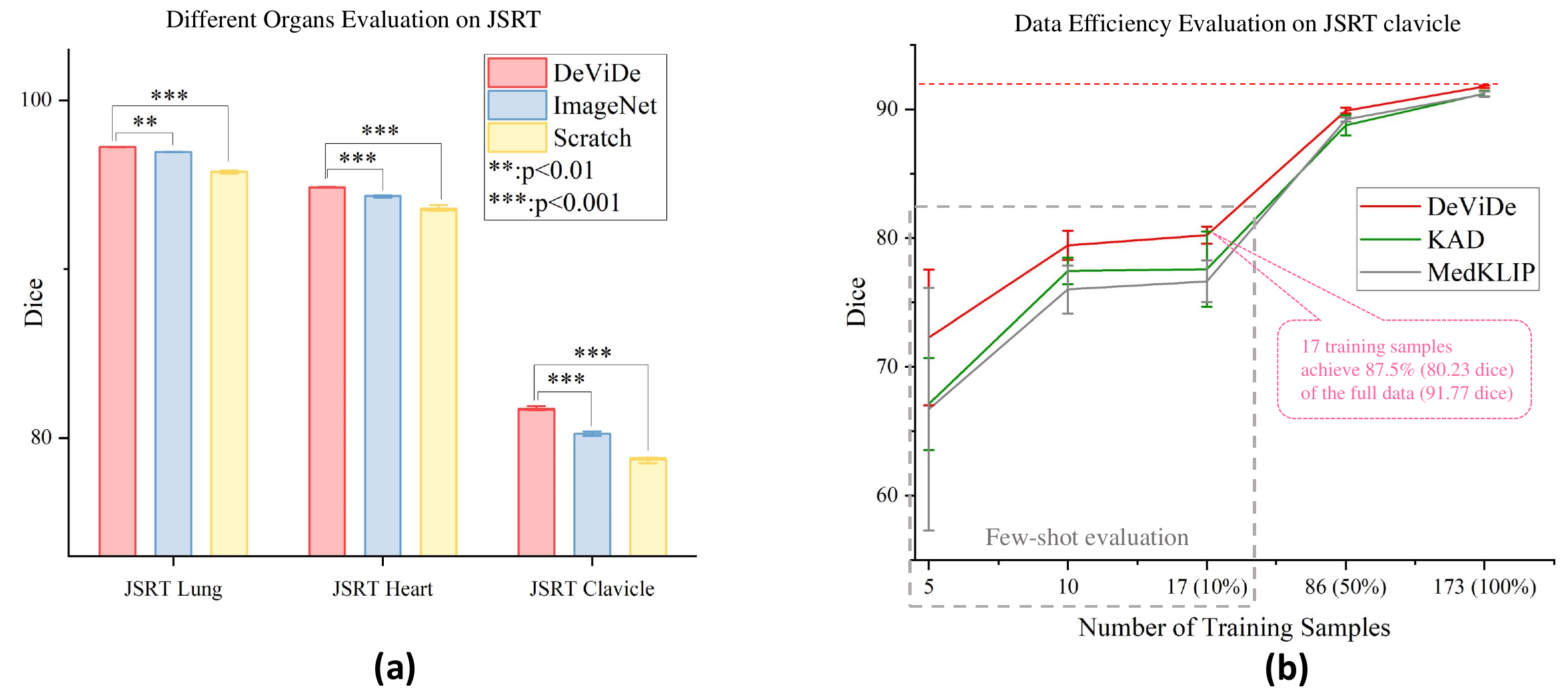}
    \caption{Evaluation of segmentation ability across organs and few-shot segmentation under full fine-tuning setting. (a) The bar chart compares Dice coefficients of different training approaches for lung, heart, and clavicle segmentation using the JSRT dataset. Our method significantly outperforms others, with the clavicle segmentation task gaining the most. (b) The line graph assesses data efficiency in few-shot learning, focusing on the JSRT clavicle dataset. DeViDe achieves a Dice coefficient close to full dataset performance with just 17 training samples, highlighting its few-shot learning capability.}
    \label{fig:JSRT_seg}
\end{figure}

Focusing on the JSRT dataset (c.f. Fig.~\ref{fig:JSRT_seg}a) with multi-organ segmentation, we observe a statistically significant improvement in Dice score for the lung, heart, and clavicle. It's worth noting that DeViDe offers the highest performance gain on the clavicle, which is the most challenging class to segment among the three.
 
We extend the challenging clavicle segmentation task to a few-shot scenario to assess the advantages of data efficiency offered by pre-training methods. As depicted in Fig.~\ref{fig:JSRT_seg}b, fine-tuning based on DeViDe segmentation significantly outperforms KAD and MedKLIP by a wide margin in low-data regimes, gradually reducing the margin when all the data is utilized, while maintaining its position as the top performer.

\begin{table}[!t]
        \centering

        \caption{Comparison with fully-supervised pre-trained model on different downstream segmentation tasks via Dice score.}
        \label{tab: fully-supervised} 
        \begin{tabular}{cccccc}
        \toprule    
             \multirow{2}*{Backbone} & Pretraining data & \multirow{2}*{ChexDet}  & \multirow{2}*{SIIM} & \multirow{2}*{Montgomery}  \\
              & and methods& & & &  \\
        \midrule
            \multirow{3}*{ViT-B} & Random & 57.78 $\pm$ 0.24 & 72.21 $\pm$ 1.70 & 92.03 $\pm$ 0.20\\
            & ImageNet-1K & 69.70 $\pm$ 0.17 & 73.06 $\pm$ 0.76 & 96.59 $\pm$ 0.11\\
            & \textbf{DeViDe} & \textbf{70.27$\pm$0.08} & \textbf{75.52$\pm$0.46} & \textbf{96.75$\pm$0.08}\\
        \bottomrule
        \end{tabular}
    \end{table}

\subsection{Ablation on the losses}
\begin{figure}[htbp]
    \centering
    \subfloat[]{\includegraphics[width=0.50\textwidth]{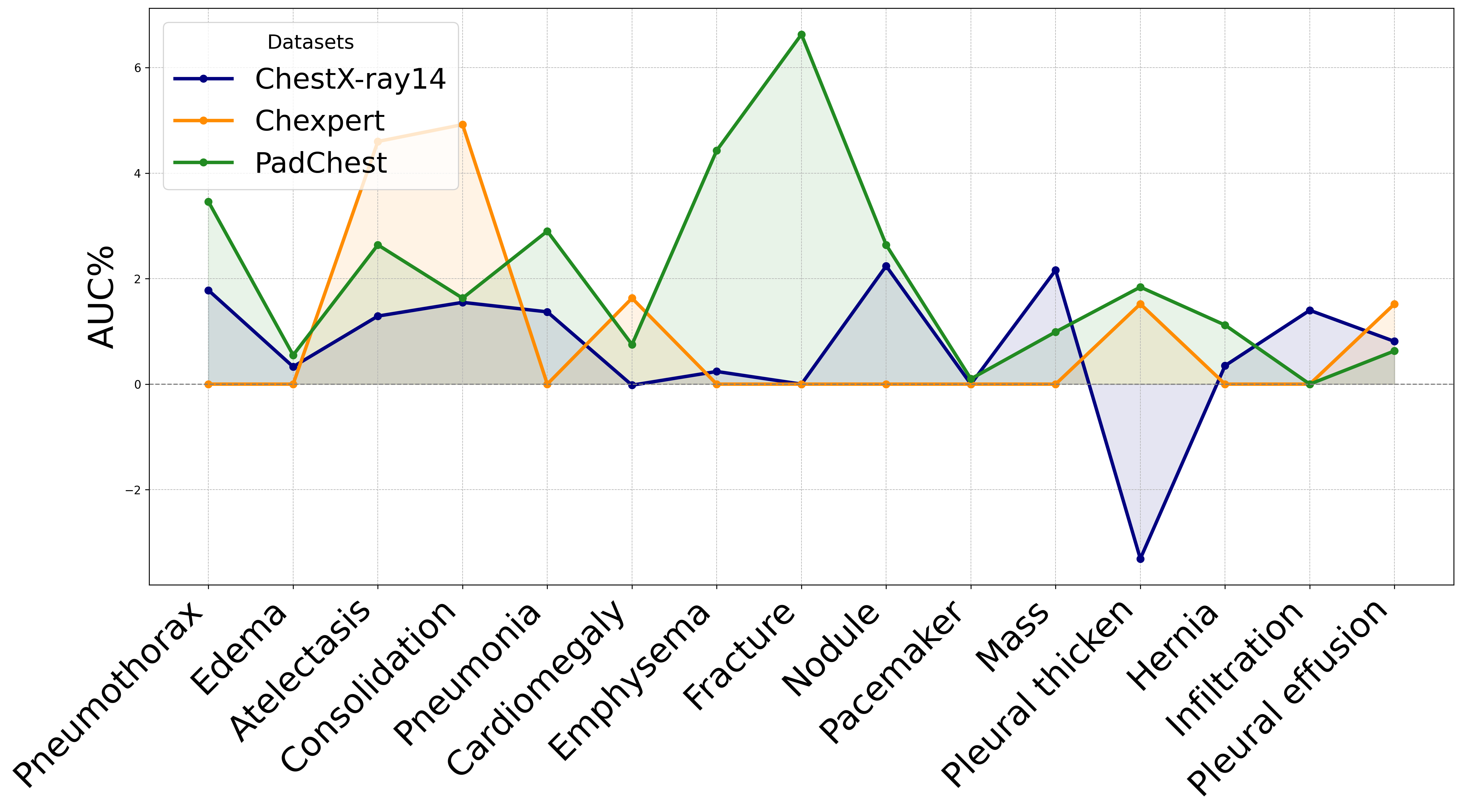}}
    \hfill
    \tiny
    \subfloat[]{
    \begin{tabular}{lccc}
    \toprule
    & \textbf{ChestX-ray14} & \textbf{Chexpert} & \textbf{PadChest} \\
    \midrule
    Pneumothorax & 1.78\% & -- & 3.46\%\\
    Edema & 0.33\%  & 4.60\% & 0.55\% \\
    Atelectasis & 1.29\%  & 4.60\%  & 2.64\%  \\
    Consolidation & 1.55\%  & 4.92\%  & 1.63\%  \\
    Pneumonia & 1.37\%  & -- & 2.90\%  \\
    Cardiomegaly & \textcolor{gray}{-0.02}\%  & 1.63\% & 0.75\%  \\
    Emphysema & 0.24\%  & -- & 4.43\% \\
    Fracture & -- & -- & 6.63\%  \\
    Nodule & 2.24\% & -- & 2.64\% \\
    Pacemaker & -- & -- & 0.10\% \\
    Mass & 2.16\% & -- & 0.99\%  \\
    Pleural thicken & \textcolor{gray}{-3.31}\% & -- & 1.84\%  \\
    Hernia & 0.35\% & -- & 1.12\%  \\
    Infiltration & 1.40\% & -- & --  \\
    Pleural effusion & 0.81\% & 1.52\%  & 0.63\% \\
    
    \bottomrule
  \end{tabular}
    }
    \caption{Evaluation of radiologic descriptions' impact on zero-shot inference. (a) Plot shows AUC score enhancements for various pathologies post knowledge injection using descriptions, (b) Corresponding data table lists specific AUC improvements for pathologies within multiple datasets.}
    \label{image:ablations}
\end{figure}

We conducted an ablation to study the contribution of the tough-negative sampling ($\mathcal{L}_{tnc}$, Eq.~\ref{eq:l_tnc}) as well as that of the radiographic (visual) descriptions of the entities ($\mathcal{L}_{pta}$, Eq.~\ref{eq:l_pta}). As recorded Table.~\ref{table:ablations}, we repeat the ZSL-setting on the three classification datasets without $\mathcal{L}_{tnc}$ as well as $\mathcal{L}_{pta}$ and observe a drop in performance. For example, the classification accuracy drops by more than 3\% on two datasets without local patch-text alignment. Furthermore, we explored how the injection of visual description of radiology entities enhances  the zero-shot inference.  In  Fig.~\ref{image:ablations}, we present an entity-wise `performance gain' contributed by $\mathcal{L}_{pta}$ across the entities common among the three test datasets and the relevant entity set $\mathcal{E}$, which are augmented by the description. Except in two instances of ChestX-ray14, we see that the image-to-description alignment consistently provides a positive performance boost, in some cases up to 6.6\% in AUC (Fracture class on PadChest).


\begin{table}[!t]
\centering

\caption{Ablation study delineating the impact on the zero-shot learning capability of DeViDe without TNC and PTA losses.}
\label{table:ablations}
\captionsetup{skip=2pt}
\begin{tabular}{lccccccc}
\hline
\multirow{2}*{Methods}& \multicolumn{3}{c}{ChestXray14} & \multicolumn{3}{c}{CheXpert} & \multicolumn{1}{c}{PadChest} \\ 
\cmidrule(r){2-4} \cmidrule(l){5-7} \cmidrule(l){8-8}
 & AUC & MCC & ACC & AUC & MCC & ACC & AUC \\ 
\hline
\textbf{DeViDe} & \textbf{0.7771} & \textbf{0.2699} & 0.8228 & \textbf{0.8998} & \textbf{0.5660} & \textbf{0.8825} & \textbf{0.7465}  \\
w/o $\mathcal{L}_{pta}$ & 0.7730 & 0.2571 & 0.7874 & 0.8965 & 0.5364 & 0.8538 & 0.7397  \\
w/o $\mathcal{L}_{pta}$ and $\mathcal{L}_{tnc}$ & 0.7689 & 0.2614 & \textbf{0.8250} & 0.8707 & 0.5163 & 0.8663 &  0.7348  \\ 
\hline
\end{tabular}
\end{table}

\subsection{Qualitative Analysis}

\begin{figure}[t!]

    \centering
    \includegraphics[width=12.5cm]{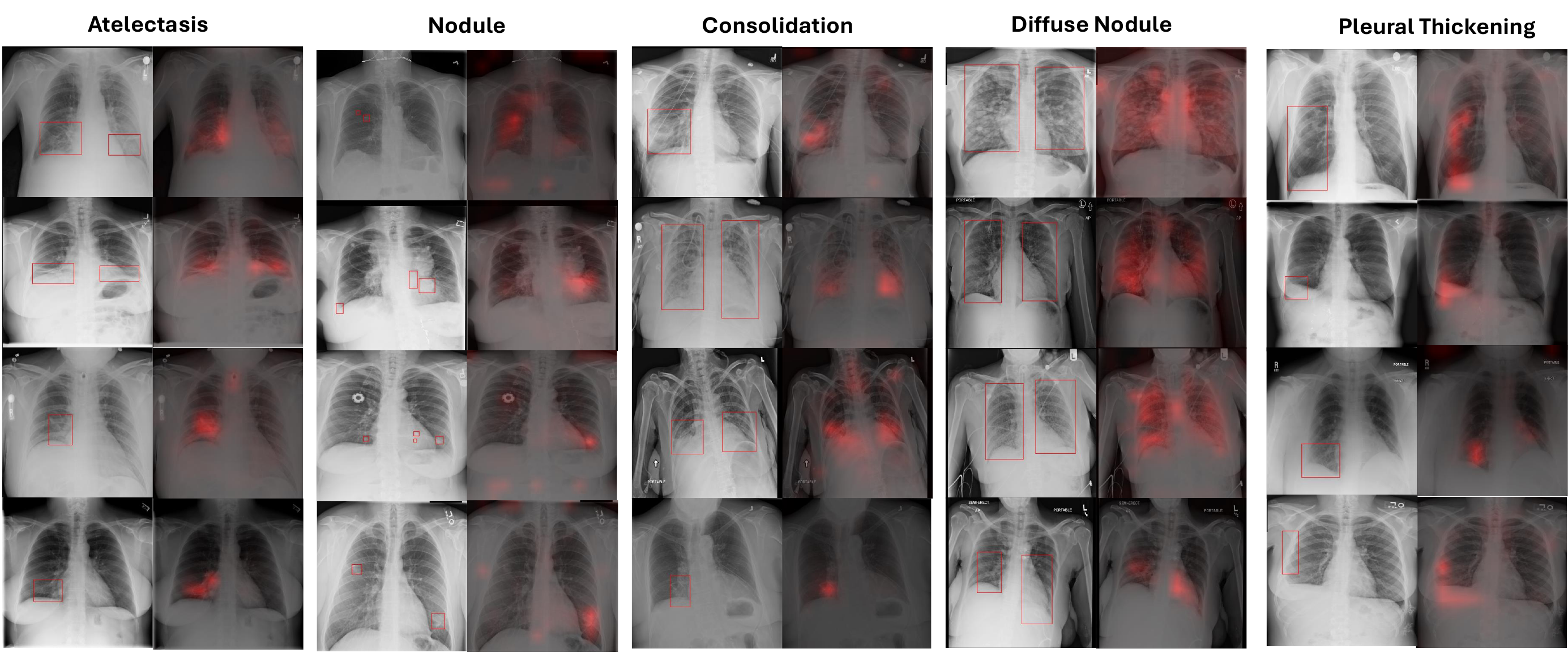}
    \caption{Zero-shot visual grounding of diseases in findings using samples from the ChestX-Det dataset. It's noteworthy that the attention for `diffuse nodule' is dispersed, whereas the attention on `nodule' is concentrated.}
    \label{image:grounding}
\end{figure}

\paragraph{Visual grounding of diseases.} Vision-language models offer inherent explainability in terms of visual grounding of the findings in the radiographs, thanks to the cross-attention layer. In Fig.\ref{image:grounding}, we illustrate this grounding on the test set of CheX-Det dataset, which also consists of bounding boxes for the abnormal regions. Specifically, we plot the attention mask of the knowledge retrieval layer as employed during inference with the class name as the text input. Observe the perfect localization of the abnormalities is the diverse set of examples. Interestingly, observe the diffused nature of attention for `Diffuse Nodule' and the localized attention for `Nodule', which highlights the strength of granular training.    

\begin{figure}[t!]

    \centering
    \includegraphics[width=\linewidth]{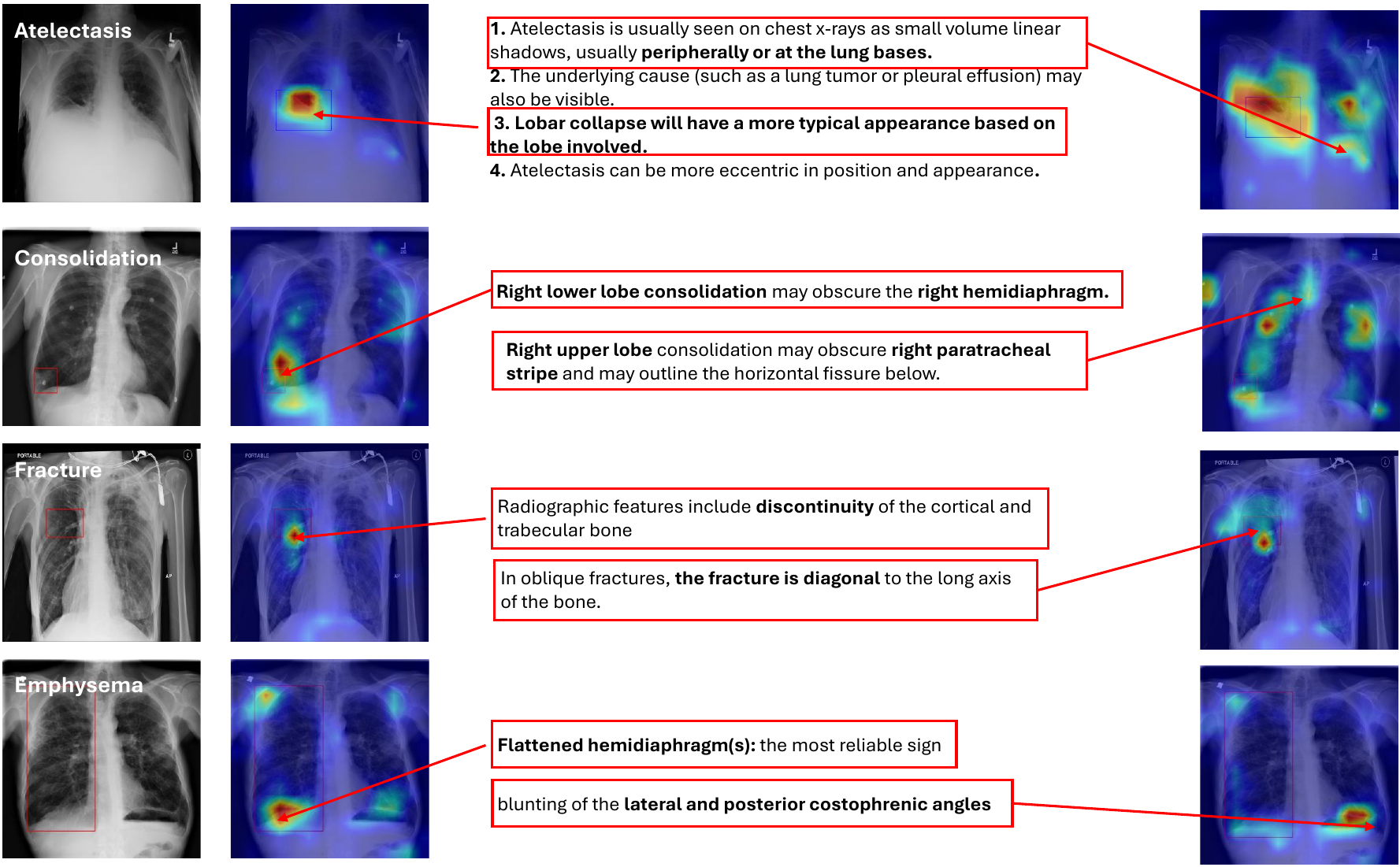}
    \caption{Image-to-description matching by sentence-level visual attention on images randomly chosen samples from ChestX-Det dataset. Observe that the attention corresponds to the sentence while being complementary to the finding class.}
    \label{image:sentenceattention}
\end{figure}

\paragraph{Image-to-description matching.} Finally, we check how the descriptions attend to the image regions. Similar to visual grounding, we extract attention from the knowledge-retrieval layer. However, now the encoded text is the visual descriptions of the findings, as shown in Fig.\ref{image:sentenceattention}. Notice the correspondence between the level of detail in the description to the broadness of the heatmap, for e.g. With Atelectasis, the attention to lungs is broader while that to lobes is narrower. Therefore, the description provides supplementary knowledge from which discriminative features can eventually be inferred.

\section{Conclusion}
In this study, we present DeViDe, an novel Vision-Language Pre-training (VLP) model that enhances medical pre-training with multi-faceted knowledge: reports, medical definitions, and visual descriptions of radiology entities. We encode these facets into the model by guiding its training using tailored loss objectives aligning for global and local alignment. DeViDe has undergone thorough evaluation in a zero-shot setting, where is showed a significantly superior performance relative to prior work on three datasets. When fine-tuned with DeViDe's initialization, both classification and segmentation tasks converged to a better optimum in terms of their performance. In a few-shot setting, DeViDe-initialization was even shown to be more data efficient. Finally, through ablation studies and qualitative analysis, we indicate that the contribution of detailed radiographic visual descriptions is paramount in this performance gain.
\\

\bibliographystyle{splncs04}
\bibliography{egbib}
\end{document}



\title{Supplementary Material} 

\titlerunning{Abbreviated paper title}

\author{First Author\inst{1}\orcidlink{0000-1111-2222-3333} \and
Second Author\inst{2,3}\orcidlink{1111-2222-3333-4444} \and
Third Author\inst{3}\orcidlink{2222--3333-4444-5555}}

\authorrunning{F.~Author et al.}

\institute{Princeton University, Princeton NJ 08544, USA \and
Springer Heidelberg, Tiergartenstr.~17, 69121 Heidelberg, Germany
\email{lncs@springer.com}\\
\url{http://www.springer.com/gp/computer-science/lncs} \and
ABC Institute, Rupert-Karls-University Heidelberg, Heidelberg, Germany\\
\email{\{abc,lncs\}@uni-heidelberg.de}}

\maketitle

We present the supplementary material in three sections: In Sec.~\ref{sec:datasets}, we provide a detailed overview of the datasets employed in our work along with the tasks they have been employed for. In Sec.~\ref{sec:grouding}, more qualitative and quantitative analysis of zero-shot grounding task are provided. In Sec.~\ref{sec:disease_details}, we detail the disease-by-disease performance of ChestX-ray14 dataset and CheXpert dataset.

\section{Target Tasks and Datasets}
\label{sec:datasets}

We pretrain our DeViDe models on MIMIC-CXR~\cite{johnson2019mimic} dataset and evaluate by zero-shot and finetuning on five classification datasets CheXpert~\cite{irvin2019chexpert}, ChestX-ray14~\cite{wang2017chestx}, PadChest~\cite{bustos2020padchest}, NIH Shenzhen CXR~\cite{jaeger2014two}, RSNA Pneumonia~\cite{rsna}, two organ segmentation datasets JSRT~\cite{shiraishi2000development} Montgomery~\cite{jaeger2014two} and two disease segmentation ChestX-Det~\cite{lian2021structure}, SIIM-ACR~\cite{siim-acr}. The details are shown in Tab. \ref{tab:datasets} and below: 

\begin{table}[t]
\centering
\caption{DeViDe is pre-trained and validated on ten public CXR datasets. To evaluate the performance of our method, we implement zero-shot as well as finetuning and data efficiency evaluations on classification, segmentation tasks.}
\label{tab:datasets}
\scriptsize
\begin{tabular}{lllll}\toprule
Dataset &Task &Usage*&(Pre)train/val/test \\\midrule
MIMIC-CXR & classify 14 thoracic diagnoses & P & 377110/*/*\\
CheXpert & classify 14 thoracic diagnoses & F|Z & 223414/234/668\\
ChestX-ray14&classify 14 thoracic diseases &F|Z &75312/11212/25596\\
PadChest & classify 19 thoracic diagnoses & Z & */*/39700\\
RSNA Pneumonia &classify lung opacity, abnormality &F &21295/2680/2709\\
Shenzhen &classify tuberculosis &F &463/65/134\\
Montgomery & segment lungs & F & 92/15/31 \\
JSRT&segment lungs, heart, clavicles &F|E &173/25/49\\
ChestX-Det& segment 13 thoracic diseases  & F& 2722/303/553\\
SIIM-ACR &segment pneumothorax &F &9607/1068/1372\\
\bottomrule
\end{tabular}
\begin{tablenotes} 
\item * The usage of each dataset in our experiments is denoted with P for pretraining, F for fine-tuning, Z for zero-shot inference and E for data efficiency evaluation.

\end{tablenotes}
\end{table}

\begin{itemize}
    \item {\bf MIMIC-CXR}~\cite{johnson2019mimic}, a public dataset of chest radiographs with radiology text reports, which contains 377,110 images corresponding to 227,835 radiographic studies for 65,379 patients. We follow the data preprocess of~\cite{zhang2023knowledge} and pretrain on the whole images and their corresponding texts.
    \item {\bf CheXpert}~\cite{irvin2019chexpert}, which includes 224K chest radiographs of 65240 patients and capturing 14 thoracic diseases. For finetuning we train on the 14 thoracic diseases and use the official training data split 224K for training, validation set 234 images and test set 668 images. We follow the standard practice by testing on 5 diseases (Atelectasis, Cardiomegaly, Consolidation, Edema, and Pleural Effusion).
    \item {\bf ChestX-ray14}~\cite{wang2017chestx}, which contains 112K frontal-view X-ray images of 30805 unique patients with the text-mined fourteen disease image labels (where each image can have multi-labels). We use the official training set 86K (90\% for training and 10\% for validation) and testing set 25K. The downstream models are trained to predict 14 pathologies in a multi-label classification setting and the mean AUC score is utilized to evaluate the classification performance. We use its test set for zero-shot and the metric evaluations are AUC, F1 and ACC. 
    \item {\bf PadChest}~\cite{bustos2020padchest}, includes more than 160,000 images obtained from 67,000 patients with 19 differential diagnoses. We utilize the same data split with~\cite{zhang2023knowledge} and implement zero-shot evaluation on 39k test images.
    \item {\bf NIH Shenzhen CXR}~\cite{jaeger2014two}, which contains 326 normal and 336 Tuberculosis (TB) frontal-view chest X-ray images. We split 70\% of the dataset for training, 10\% for validation and 20\% for testing which are the same with ~\cite{ma2022benchmarking};
    \item {\bf RSNA Pneumonia}~\cite{rsna}, which consists of 26.7K frontal view chest X-ray images and each image is labeled with a distinct diagnosis, such as Normal, Lung Opacity and Not Normal (other diseases). 80\% of the images are used to train, 10\% to valid and 10\% to test. These target datasets are composed of both multi-label and multi-class classification tasks with various diseases.
    \item \textbf{JSRT}~\cite{shiraishi2000development}, which is an organ segmentation dataset including 247 frontal view chest X-ray images. All of them are in 2048$\times$2048 resolution with 12-bit gray-scale levels. Both lung, heart and clavicle segmentation masks are available for this dataset. We split 173 images for training, 25 for validation and 49 for testing.
     \item \textbf{ChestX-Det}~\cite{lian2021structure}, which is a disease segmentation dataset and an improved version of ChestX-Det10~\cite{liu2020chestx}. This dataset contains 3,578 images with instance-level annotations for 13 common thoracic pathology categories, sourced from the NIH ChestX-ray14 dataset. Annotations were provided by three board-certified radiologists, and the dataset includes additional segmentation annotations. We consolidated all the diseases into one region and the goal of segmenting this dataset is to distinguish between diseased and non-diseased areas for each image. There are official split for training and testing sets and we split 10\% images from training set for validation.
    \item \textbf{SIIM-ACR}~\cite{siim-acr}, a dataset resulting from a collaboration between SIIM, ACR, STR, and MD.ai, contains 12,089 chest X-ray images. It is the largest public pneumothorax segmentation dataset to date, comprising 3,576 pneumothorax images and 9,420 non-pneumothorax images, all of which are available in 1024x1024 pixel resolution. We randomly divided the dataset into training (80\%), validation (10\%) and testing (10\%). The segmentation performance is measured by the mean Dice which average the dice of pneumothorax non-pneumothorax images.
    \item \textbf{Montgomery}~\cite{jaeger2014two}, a dataset for lung segmentation which is acquired from the tuberculosis control program of Montgomery County's Department of Health and Human Services, MD, USA. The dataset includes 138 posterior-anterior chest X-rays, of which 80 are normal and 58 show manifestations of tuberculosis. We randomly divided the dataset into a training set (70\%), a validation set (10\%) and a test set (20\%).
\end{itemize}

\section{Grounding Tasks Performance}
\label{sec:grouding}

\begin{figure}[t!]

    \centering
    \includegraphics[width=\linewidth]{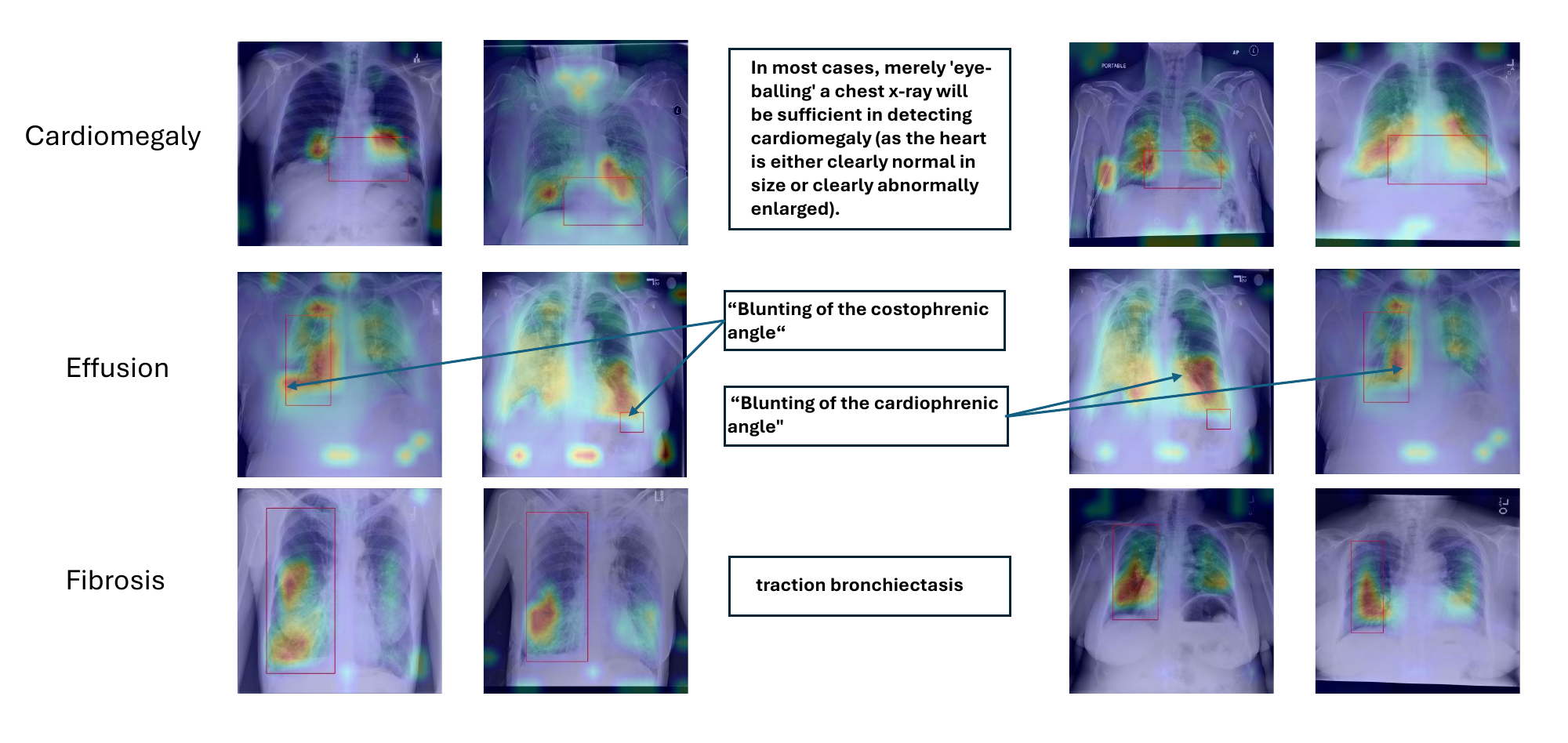}
    \caption{In the ChestX-Det dataset, when analyzing sentence-image attention of DeViDe for various diseases, it's observed that the attentional focus shifts based on the specific terms mentioned in the sentences. For example, when terms such as \textbf{"costophrenic angle"} and \textbf{"cardiophrenic angle"} are included, the model's attention notably moves towards the relevant areas on the images. This demonstrates the model's capability to dynamically adjust its focus depending on the textual cues provided.}
    \label{image:sentenceattention2}
\end{figure}

\begin{figure}[t!]
    \centering
    \includegraphics[width=1\linewidth]{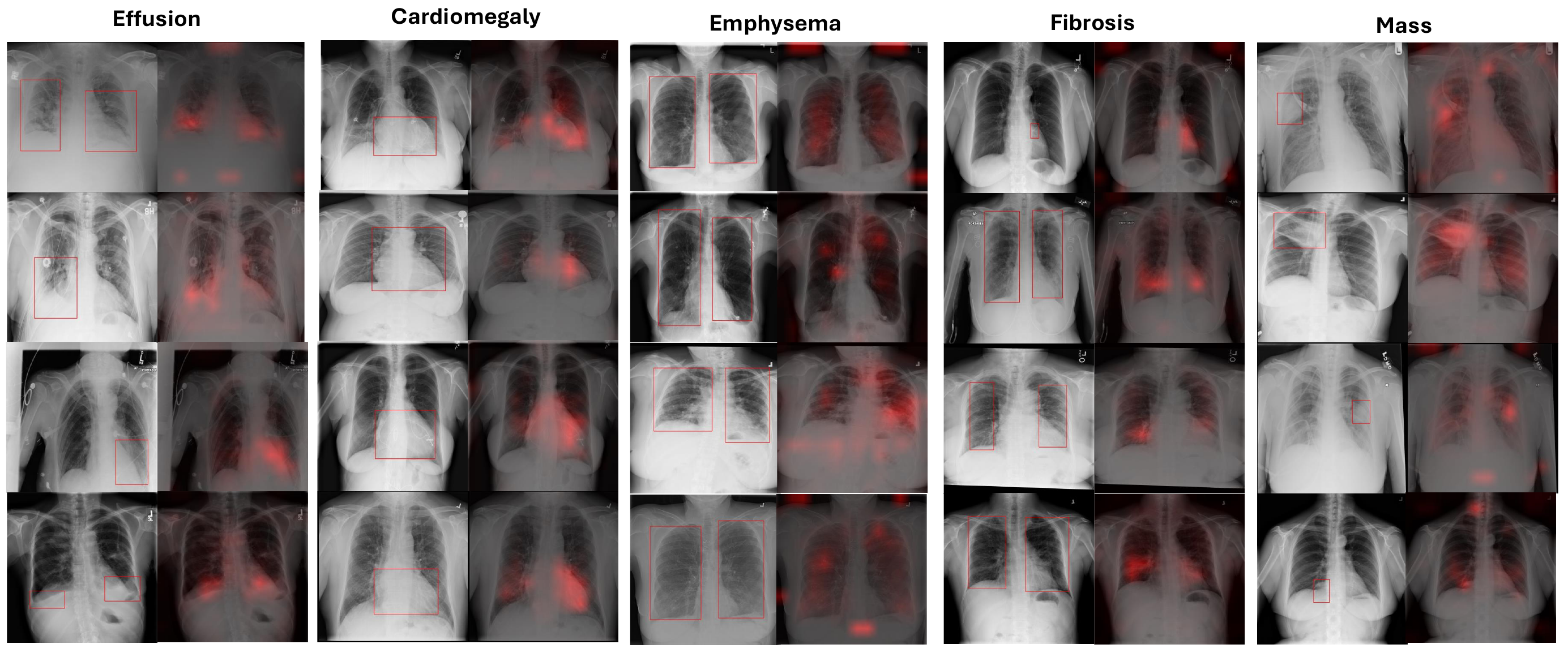}
    \caption{In further instances of zero-shot visual grounding for a variety of diseases using samples from the ChestX-Det dataset, our approach remains effective even when there are multiple focal points in an image. Specifically, in cases such as Effusion, our method maintains complete attention on different disease-related spots.}
    \label{fig:grounding_2}
\end{figure}

In this section, we further analyze, quantatively and qualitatively, the grounding abilities of DeViDe in a zero-shot setting. Specifically, we provide qualitative examples for visual grounding in a wider set of diseases. Fig.~\ref{image:sentenceattention2} shows the sentence level attention computed over the radiological descriptions and \ref{fig:grounding_2} illustrates the label-level grounding. 

\begin{figure}[t!]
    \centering
    \includegraphics[width=1\linewidth]{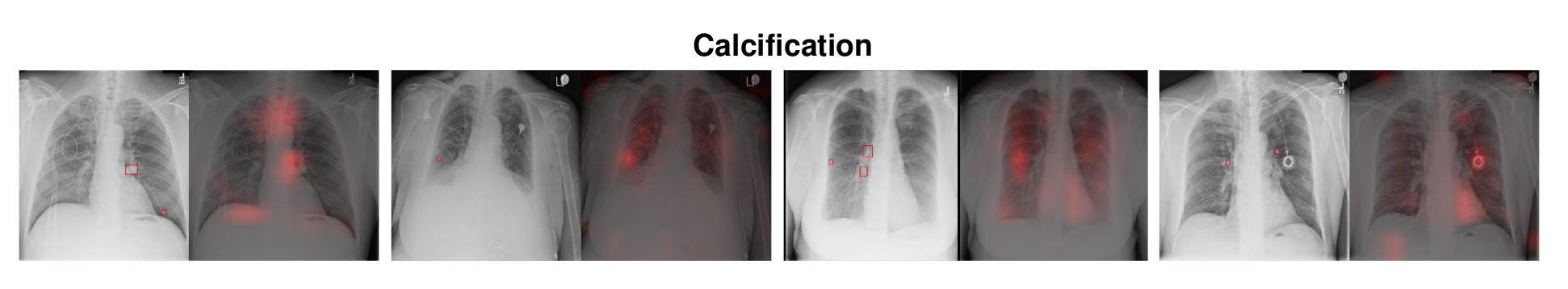}
    \caption{In the qualitative analysis of Calcification, DeViDe demonstrated a capacity for localizing proximate image blocks, although it did not achieve complete overlap. This limitation was primarily due to the diminutive and abundant nature of the lesion points.}
    \label{fig:grounding_cali}
\end{figure}

For evaluating the grounding abilities for DeViDe quantitatively, we perform the \textbf{Pointing Game}, which evaluates how accurately a model can pin-point specific areas of interest in an image. It works by comparing the model's primary focus-area in an image to the actual region-of-interest (ground-truth). If the model's focus matches the correct area, it's a `hit'; otherwise, it's a `miss'. In Table \ref{tab:pointing_game}, when we compare DeViDe's performance in the pointing game with the baseline methods, we find that DeViDe correctly identifies the Region of Interest (RoI) better than the baselines in six out of ten cases, and it also achieves a higher average performance. We see that the mean performance takes a major hit due to the exceptionally poor performance on `Calcification', which we qualitatively analyze in Fig.\ref{fig:grounding_cali}. We see that even though DeViDe attends to the calcified region, the detection is missed due to the surrounding noise, especially owing to the small size of calcification.

\begin{table}[t]
\centering
\scriptsize
\caption{Pointing Game: DeviDe achieves the best performance in ChestX-Det10 grounding task under zero-shot setting, where underline number indicates the second best performance.}
\label{tab:pointing_game}
\begin{tabular}{l|l|l l l l l l l l l l l}
\hline
Metric         & Method   & \rotatebox{75}{Mean}  & \rotatebox{75}{Atelectasis} & \rotatebox{75}{Calcification} & \rotatebox{75}{Consolidation} & \rotatebox{75}{Effusion} & \rotatebox{75}{Emphysema} & \rotatebox{75}{Fibrosis} & \rotatebox{75}{Fracture} & \rotatebox{75}{Mass}   & \rotatebox{75}{Nodule} & \rotatebox{75}{Pneumothorax} \\ \hline
  & GLORIA   & 0.367 & 0.479       & 0.053         & \underline{0.737}         & 0.528    & \underline{0.667}     & 0.366    & 0.013    & \underline{0.533}  & 0.156   & \underline{0.143}        \\
Pointing& BioViL   & 0.380 & 0.375       & 0.105         & 0.664         & 0.615    & \textbf{0.795}     & \underline{0.378}   & 0.013    & 0.500  & 0.130   & \textbf{0.229}        \\
  Game& KAD      & \underline{0.391} & \underline{0.646}       & \textbf{0.132}         & 0.699         & \underline{0.618}    & 0.644     & 0.244    & \textbf{0.199}    & 0.267  & \textbf{0.316}   & \underline{0.143}        \\
  & DeViDe      & \textbf{0.424} & \textbf{0.667}       & 0.026         & \textbf{0.761}         & \textbf{0.656}    & 0.590     & \textbf{0.390}    & \underline{0.158}    & \textbf{0.545}  &  \underline{0.215}   & \textbf{0.229}        \\\hline
\end{tabular}
\end{table}

\section{Disease-by-Disease Performance}
\label{sec:disease_details}
In this section, we detail the performance of DeViDe on a disease-by-disease basis under zero-shot conditions on the CheXpert dataset, with 5 main findings, and on the ChestX-ray14 dataset. The results are presented in Table.\ref{tab:chexpert} and Table.\ref{tab:chestxray-14}, respectively.

\begin{table}[t]
\centering
\begin{minipage}[t]{0.45\textwidth}
\centering
\caption{Performance of DeViDe across 5 main findings in the CheXpert dataset.}
\label{tab:chexpert}
\begin{tabular}{l|l}
\hline
\textbf{Metric}    & \textbf{AUC} \\ \hline
Mean & 0.900 \\
Atelectasis & 0.878 \\
Cardiomegaly & 0.863 \\
Consolidation & 0.876 \\
Edema & 0.925 \\
Pleural Effusion & 0.957 \\ \hline
\end{tabular}
\end{minipage}\hfill
\begin{minipage}[t]{0.45\textwidth}
\centering
\caption{Performance of DeViDe across 14 disease classes in the ChestX-ray14 dataset.}
\label{tab:chestxray-14}
\begin{tabular}{l|l}
\hline
\textbf{Finding}   & \textbf{AUC} \\ \hline
Mean & 0.777 \\
Atelectasis & 0.761 \\
Cardiomegaly & 0.863 \\
Pleural Effusion & 0.817 \\
Infiltration & 0.695 \\
Lung Mass & 0.772 \\
Lung Nodule & 0.709 \\
Pneumonia & 0.701 \\
Pneumothorax & 0.849 \\
Consolidation & 0.726 \\
Edema & 0.825 \\
Emphysema & 0.866 \\
Fibrosis & 0.719 \\
Pleural thicken & 0.628 \\
Hernia & 0.936 \\ \hline
\end{tabular}
\end{minipage}
\end{table}

\bibliographystyle{splncs04}
\bibliography{egbib}